\documentclass[11pt]{article}
\usepackage{eamt20}
\usepackage{times}
\usepackage{url}
\usepackage{latexsym}
\usepackage[small,bf]{caption} 
\usepackage{graphicx}
\usepackage{todonotes}

\title{Human Parity in Machine Translation According to Humans}
\title{Reassessing Claims of Human Parity and Super-Human Performance in Machine Translation at WMT 2019}

\author{Antonio Toral\\
Center for Language and Cognition\\
University of Groningen\\
  The Netherlands\\  {\tt a.toral.ruiz@rug.nl}}

\date{}

\begin{document}
\maketitle
\begin{abstract}
We reassess the claims of human parity and super-human performance made at the news shared task of WMT 2019 for three translation directions: English$\rightarrow$German, English$\rightarrow$Russian and German$\rightarrow$English.
First we identify three potential issues in the human evaluation of that shared task:
(i) the limited amount of intersentential context available, (ii) the limited translation proficiency of the evaluators and (iii) the use of a reference translation.
We then conduct a modified evaluation taking these issues into account.
Our results indicate that all the claims of human parity and super-human performance made at WMT 2019 should be refuted, except the claim of human parity for English$\rightarrow$German.
Based on our findings, we put forward a set of recommendations and open questions for future assessments of human parity in machine translation.
\end{abstract}


\section{Introduction}

The quality of the translations produced by machine translation (MT) systems has improved considerably since the adoption of architectures based on neural networks~\cite{D16-1025}. 
To the extent that, in the last two years, there have been claims of MT systems reaching human parity and even super-human performance~\cite{achieving-human-parity-on-automatic-chinese-to-english-news-translation,bojar-EtAl:2018:WMT1,barrault-etal-2019-findings}.
Following Hassan et al.~\shortcite{achieving-human-parity-on-automatic-chinese-to-english-news-translation}, we
consider that human parity is achieved for a given task {\it t} if the performance attained by a computer on {\it t} is equivalent to that of a human, i.e. there is no significant difference between the performance obtained by human and by machine.
Super-human performance is achieved for {\it t} if the performance achieved by a computer is significantly better than that of a human.


Two claims of human parity in MT were reported in 2018.
One by Microsoft, on news translation for Chinese$\rightarrow$English~\cite{achieving-human-parity-on-automatic-chinese-to-english-news-translation}, and another at the news translation task of WMT for English$\rightarrow$Czech~\cite{bojar-EtAl:2018:WMT1}, in which MT systems Uedin~\cite{haddow-EtAl:2018:WMT} and Cuni-Transformer~\cite{kocmi-sudarikov-bojar:2018:WMT} reached human parity and super-human performance, respectively.
In 2019 there were additional claims at the news translation task of WMT~\cite{barrault-etal-2019-findings}: human parity for German$\rightarrow$English, by several of the submitted systems, and for English$\rightarrow$Russian, by system Facebook-FAIR~\cite{ng-EtAl:2019:WMT}, as well as super-human performance for English$\rightarrow$German, again by 
Facebook-FAIR.

The claims of human parity and super-human performance in MT made in 2018~\cite{achieving-human-parity-on-automatic-chinese-to-english-news-translation,bojar-EtAl:2018:WMT1} have been since refuted given three issues in their evaluation setups \cite{laeubli2018parity,toral-EtAl:2018:WMT}: 
(i) part of the source text of the test set was not original text but translationese, (ii) the sentences were evaluated in isolation, and (iii) the evaluation was not conducted by translators.
However, the evaluation setup of WMT 2019 was modified to address some of these issues: the first issue (translationese) was fully addressed, while the second (sentences evaluated in isolation) was partially addressed, as we will motivate in Section~\ref{s:issues_context}, whereas the third (human evaluation conducted by non-translators) was not acted upon.
Given that some of the issues that led to refute the claims of human parity in MT made in 2018 have been addressed in the set-up of the experiments leading to the claims made in 2019, but that some of the issues still remain,
we reassess these later claims.


The remainder of this paper is organised as follows.
Section~\ref{s:issues} discusses 
the potential issues in the setup of the human evaluation at WMT 2019. 
Next, in Section~\ref{s:evaluation} we conduct a modified evaluation of the MT systems that reached human parity or super-human performance at WMT 2019. 
Finally, 
Section~\ref{s:disc_fw} presents our conclusions and recommendations. 

\section{Potential 
Issues in the Human Evaluation of WMT 2019}\label{s:issues}

This section discusses 
the potential 
issues that we have identified in the human evaluation of the news translation task at WMT 2019, and motivates why they might have had contributed to the fact that some of the systems evaluated therein reached human parity or super-human performance.
These issues concern the limited amount of intersentential context provided to the evaluators (Section~\ref{s:issues_context}), the fact that the evaluations were not conducted by translators (Section~\ref{s:issues_evaluators}) and the fact that the evaluation was reference-based for one of the translation directions (Section~\ref{s:issues_refbased}).

\subsection{Limited Intersentential Context}\label{s:issues_context}

In the human evaluation at previous editions of WMT evaluators had no access to intersentential context since the sentences were shown to evaluators in random order. That changed in
WMT 2019~\cite{barrault-etal-2019-findings}, which had two evaluation settings that contained intersentential context:


\begin{figure}[htbp]
 \centering
 \includegraphics[width=0.5\textwidth, trim = {0 0.5cm 0 0.5cm}]{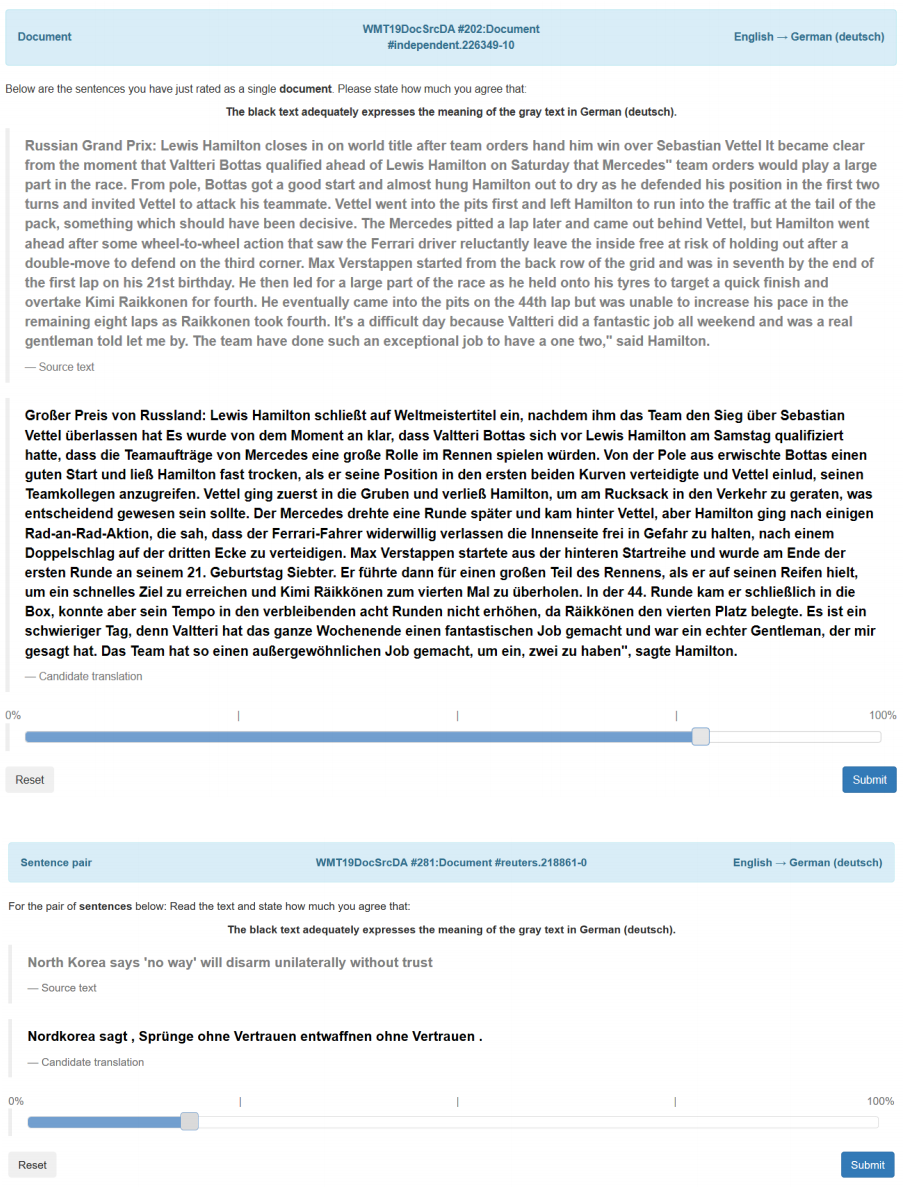}
  \caption{A snapshot of an assessment using setting DR+DC (top) and SR+DC (bottom) at WMT 2019, taken from Barrault et al.~\shortcite{barrault-etal-2019-findings}}
  \label{f:drdc_srdc}
\end{figure}

%

\begin{itemize}\itemsep0em
\item Document-level (DR+DC), inspired by L{\"a}ubli et al.~\shortcite{laeubli2018parity}, in which the whole document is available and it is evaluated globally (see top of Figure \ref{f:drdc_srdc}). 
 While the evaluator has access to the whole document, this set-up has the drawback of resulting in very few ratings (one per document) and hence suffers from low statistical power~\cite{graham2019translationese}.
\item Sentence-by-sentence with document context (SR+DC), in which segments are provided in the ``natural order as they appear in the document'' and they are assessed individually 
(see bottom of Figure \ref{f:drdc_srdc}).
Such a set-up results in a much higher number of ratings compared to the previous evaluation setting (DR+DC): one per sentence rather than one per document.
The problem with the current setting is that the evaluator can access limited intersentential context since only the current sentence is shown. This poses two issues, with respect to previous and following sentences in the document being evaluated.
With respect to previous sentences, while the evaluator has seen them recently, he/she might have forgotten some details of a previous sentence that are relevant for the evaluation of the current sentence, e.g. in long documents.
As for following sentences, the evaluator does not have access to them while evaluating the current sentence, which may be useful in some cases, e.g. when evaluating the first sentence of a document, i.e. the title of the newstory, since in some cases this may present an ambiguity for which having access to subsequent sentences could be useful.
\end{itemize}

SR+DC was the set-up used for the official rankings of WMT 2019, from which the claims of human parity and super-human performance were derived.
The requirement of information from both previous and following sentences in human evaluation of MT has 
been empirically proven in contemporary research~\cite{castilho_lrec2020}.

In our evaluation setup, evaluators are shown local context (the source sentences immediately preceding and following the current one) and are 
provided with global context: the whole source document as a separate 
text file.
Evaluators are told to 
use the global context if the local context does not provide enough information to evaluate a sentence.
In addition, evaluators are asked to evaluate all the sentences of a document in a single session.

\subsection{Proficiency of the Evaluators}\label{s:issues_evaluators}

The human evaluation of WMT 2019 was conducted by crowd workers and by MT researchers. The first type of evaluators provided roughly two thirds of the judgments (487,674) while the second type contributed the remaining one third (242,424).
Of the judgments provided by crowd workers, around half of them (224,046) were by ``workers who passed quality control''.

The fact that the evaluation was not conducted by translators might be problematic since it has been found that crowd workers lack knowledge of translation and, compared to professional translators, tend to be more accepting of (subtle) translation errors~\cite{Castil-MTSummit2017}.


Taking this into account, we will reassess the translations of the systems that achieved human parity or super-human performance at WMT 2019 with 
translators and non-translators.
The latter are 
native speakers of the target language who are not translators but who have an advanced level of the source language (at least C1 in the 
Common European Framework of Reference for Languages).


\subsection{Reference-based Evaluation}\label{s:issues_refbased}

While for two of the translation directions for which there were claims of human parity at WMT 2019 the human evaluation was reference-free (from English to both German and Russian), for the remaining translation direction for which there was a claim of parity (German to English), the human evaluation was reference-based.
In a reference-free evaluation, the evaluator assesses the quality of a translation with respect to the source sentence. Hence evaluators need to be proficient in both the source and target languages.
Differently, in a reference-based evaluation, the evaluator assesses a translation with respect, not (only) to the source sentence, but (also) to a reference translation. 

The advantage of a reference-based evaluation is that it can be carried out by monolingual speakers, since only proficiency in the target language is required.
However, the 
dependence on reference translations in this type of evaluation can lead to reference bias.
Such a bias is hypothesised to result in (i) inflated scores for candidate translations that happen to be similar to the reference translation (e.g. in terms of syntactic structure and lexical choice) and to (ii) penalise correct translations that diverge from the reference translation.
Recent research 
has found both evicence that this is the case~\cite{fomicheva-specia-2016-reference,bentivogli2018machine} and that it is not~\cite{ma-etal-2017-investigation}.


In the context of WMT 2019,
in the translation directions that followed a reference-free human evaluation, the human translation (used as reference for the automatic evaluation) could be compared to 
 MT systems in the human evaluation, just by being part of the pool of translations to be evaluated.
However, in the translation directions that followed a reference-based human evaluation, such as German$\rightarrow$English, the reference translation could not be evaluated 
against the MT systems, since it was itself the gold standard. 
A second human translation was used to this end.
In a nutshell, for English$\rightarrow$German and English$\rightarrow$Russian there is one human translation, referred to as \textsc{Human}, while for German$\rightarrow$English there are two human translations, one was used as reference and the other was evaluated against the MT systems, to which we refer to as \textsc{Ref} and \textsc{Human}, respectively.

The claim of parity for German$\rightarrow$English results therefore from the fact that \textsc{Human} and the output of an MT system (Facebook-FAIR) were compared separately to a gold standard translation, \textsc{Ref}, 
and the overall ratings that they obtained were not significantly different from each other.
If there was reference bias in this case, it could be that \textsc{Human} was penalised for being different than \textsc{Ref}.
To check whether this could be the case we use BLEU~\cite{papineni-etal-2002-bleu} as a proxy to measure the similarity between all the pairs of the three relevant translations: \textsc{Ref}, \textsc{Human} and the best MT system.
Table~\ref{t:ref_bleu} shows the three pairwise scores.\footnote{We use the \texttt{multi-bleu.perl} implementation of BLEU, giving as parameters one of the translations as the reference and the other as the hypothesis. Changing the order of the parameters results in very minor variations in the 
score.}
\textsc{HUMAN} appears to be markedly different than MT and \textsc{REF}, which are more similar to each other.

\begin{table}[htbp]
\centering
\begin{tabular}{c c c} 
\hline
MT, \textsc{Ref}       & MT, \textsc{Human}    & \textsc{Ref}, \textsc{Human}\\
\hline
35.9		& 26.5			& 21.9\\
\hline
\end{tabular}
\caption{BLEU scores between pairs of three translations (\textsc{Ref}, \textsc{Human} and the best MT system) for German$\rightarrow$English at the news translation task of WMT 2019.} 
\label{t:ref_bleu}
\end{table}

These results indicate thus that \textsc{Human} could have been penalised for diverging from the reference translation \textsc{Ref},
which could have contributed to the best MT system reaching parity.
In our experiments, we will conduct a reference-free evaluation for this translation direction 
comparing this MT system to both human translations.

\section{Evaluation}\label{s:evaluation}

\begin{figure*}[htbp]
 \centering
 \includegraphics[width=0.9\textwidth]{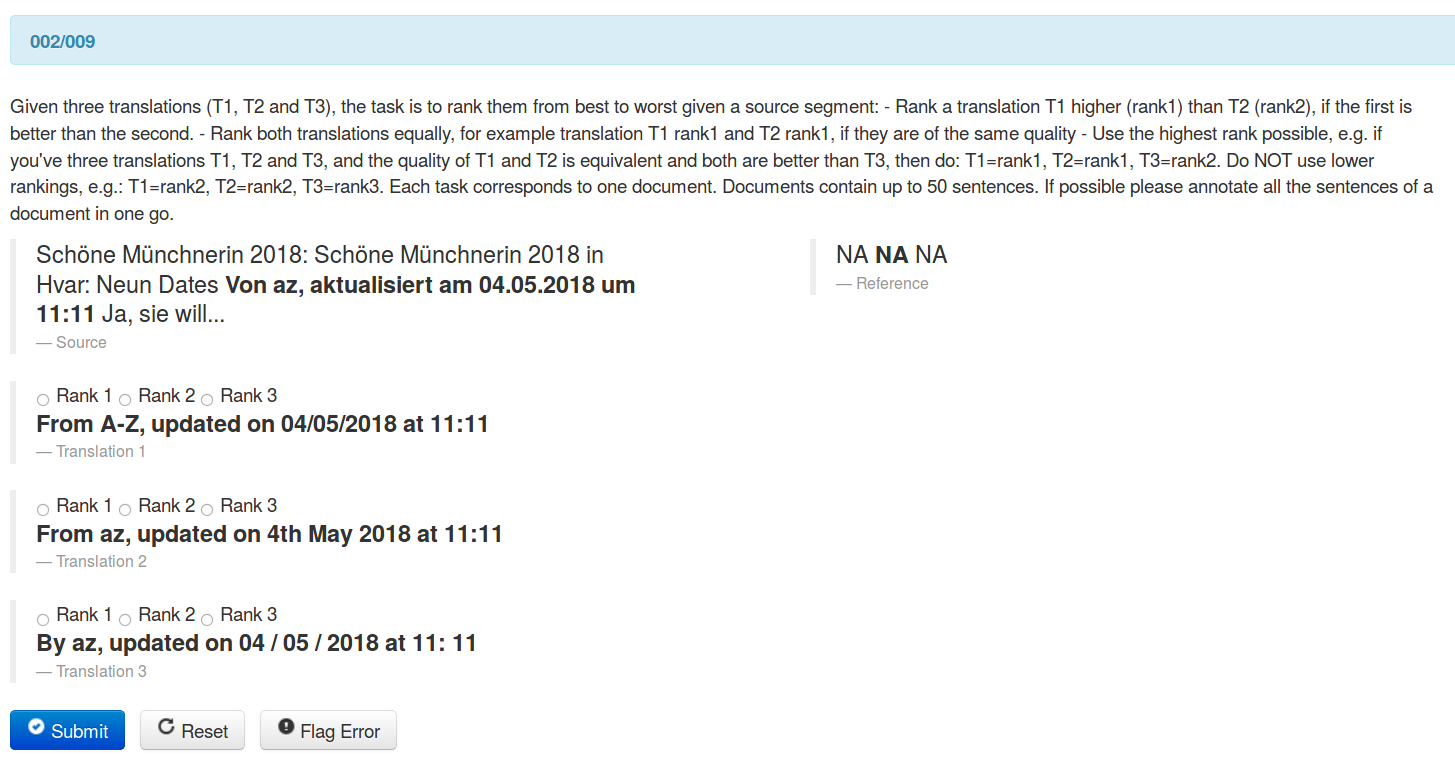}
  \caption{A snapshot of our human evaluation, for the German$\rightarrow$English translation direction, for the second segment of a document that contains nine segments. The evaluator ranks three translations, two of which are produced by human translators (\textsc{REF} and \textsc{HUMAN}) while the remaining one comes from an MT system (Facebook-FAIR), by comparing them to the source, since no reference translation is provided. Local context (immediately preceeding and following source sentences) is provided inside the evaluation tool and global context (the whole source document) is provided as a separate file.}
  \label{f:rr}
\end{figure*}

\subsection{Experimental Setup}

We conduct a human evaluation\footnote{Code and data available at \url{https://github.com/antot/human_parity_eamt2020}} for the three translation directions of WMT 2019 for which there were claims of human parity or super-human performance: German$\rightarrow$English, English$\rightarrow$German and English$\rightarrow$Russian.
We evaluate the first twenty documents of the test set for each of these language pairs. These amount to 317 sentences for German$\rightarrow$English and 302 for both English$\rightarrow$German and English$\rightarrow$Russian (the English side of the test set in all from-English translation directions is common).


We conduct our evaluation with the Appraise toolkit~\cite{mtm12_appraise}, by means of relative rankings, rather than direct assessment (DA)~\cite{graham2017can} as in Barrault et al.~\shortcite{barrault-etal-2019-findings}.
While DA has some advantages over ranking, their outcomes correlate strongly ($R>0.9$ in Bojar et al.~\shortcite{bojar-EtAl:2016:WMT1}) and the latter is more appropriate for our evaluation 
for two reasons: (i) it allows us to show the evaluator all the translations that we evaluate at once, so that they are directly compared 
(DA only shows one translation at a time, entailing that the translations evaluated are indirectly compared to each other) and (ii) it allows us to show local context to the evaluator 
(DA only shows the sentence that is being currently evaluated).

Evaluators are shown two translations for both   English$\rightarrow$German and English$\rightarrow$Russian: one by a human (referred to as \textsc{Human}) and one by the best MT system\footnote{The MT system with the highest normalised average DA score in the human evaluation of WMT 2019.} submitted to that translation direction 
(referred to as \textsc{MT}). 
For German$\rightarrow$English there are three translations (see Section~\ref{s:issues_refbased}): two by humans (\textsc{Human} and \textsc{Ref}) and one by an MT system.
The MT system is Facebook-FAIR for all three translation directions.
The order in which the translations are shown is randomised.

For each source sentence, evaluators rank the translations thereof, with ties being allowed.
Evaluators could also avoid ranking the translations of a sentence, if they detected an issue that prevented them from being able to rank them, by using the button flag error; they were instructed to do so only when strictly necessary.
Figure~\ref{f:rr} shows a snapshot of our evaluation.


From the relative rankings, we extract the number of times one of the translations is better than the other and the number of times they are tied.
Statistical significance is conducted with two-tailed sign tests, the null hypothesis being that evaluators do not prefer the human translation over MT or viceversa~\cite{laeubli2018parity}.
We report the number of successes $x$, i.e. number of ratings in favour of the human translation, and the number of trials $n$, i.e. number of all ratings except for ties.

Five evaluators took part in the evaluation for English$\rightarrow$German (two translators and three non-translators), six took part for English$\rightarrow$Russian (four translators and two non-translators) and three took part for German$\rightarrow$English (two translators and one non-translator).

Immediately after completing the evaluation, the evaluators completed a questionnaire (see Appendix~\ref{s:quest}).
It contained questions about their linguistic proficiency in the source and target languages, their amount of translation experience, the frequency with which they used the local and global contextual information, whether they thought that one of the translations was normally better than the other(s) and whether they thought that the translations were produced by human translators or MT systems.

In the remaining of this section we present the results of our evaluation for the three language pairs, followed by the inter-annotator agreement and the responses to the questionnaire.


\subsection{Results for English$\rightarrow$German}

\begin{figure}[htbp]
 \centering
 \includegraphics[width=0.5\textwidth]{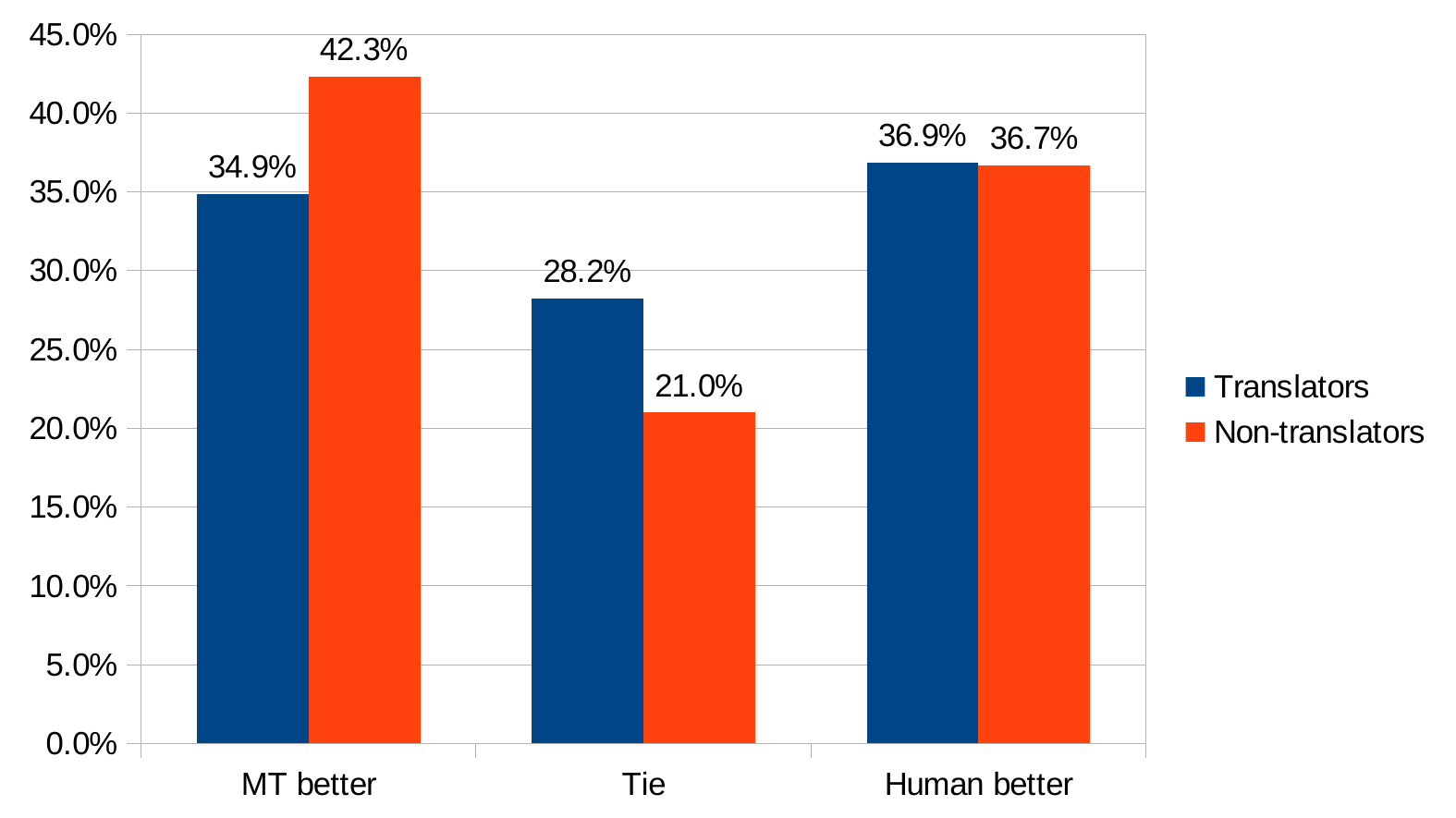}
  \caption{Results for English$\rightarrow$German for translators ($n=602$) and non-translators ($n=905$)}
  \label{f:ende}
\end{figure}

Figure~\ref{f:ende} shows the percentages of rankings\footnote{We show percentages instead of absolute numbers in order to be able to compare the rankings by translators and non-translators, as the number of translators and non-translators is not the same.} for which translators and non-translators preferred the translation by the MT system, that by the human translator or both were considered equivalent (tie).
Non-translators preferred the translation by the MT engine slightly more frequently than the human translation (42.3\% vs 36.7\%) while the opposite is observed for translators (36.9\% for \textsc{Human} vs 34.9\% for MT).
However, these differences are not significant for either translators ($x=222$, $n=432$, $p=0.6$) nor for non-translators ($x=332$, $n=715$, $p=0.06$).
In other words, according to our results there is no super-human performance, since MT is not found to be significantly better than \textsc{Human} (which was the case at WMT 2019) but \textsc{Human} is not significantly better than MT either.
Therefore our evaluation results in human parity, since the performance of the MT system and \textsc{Human} 
are not significantly different in the eyes of the translators and the non-translators that conducted the evaluation.

\begin{figure}[htbp]
 \centering
 \includegraphics[width=0.5\textwidth]{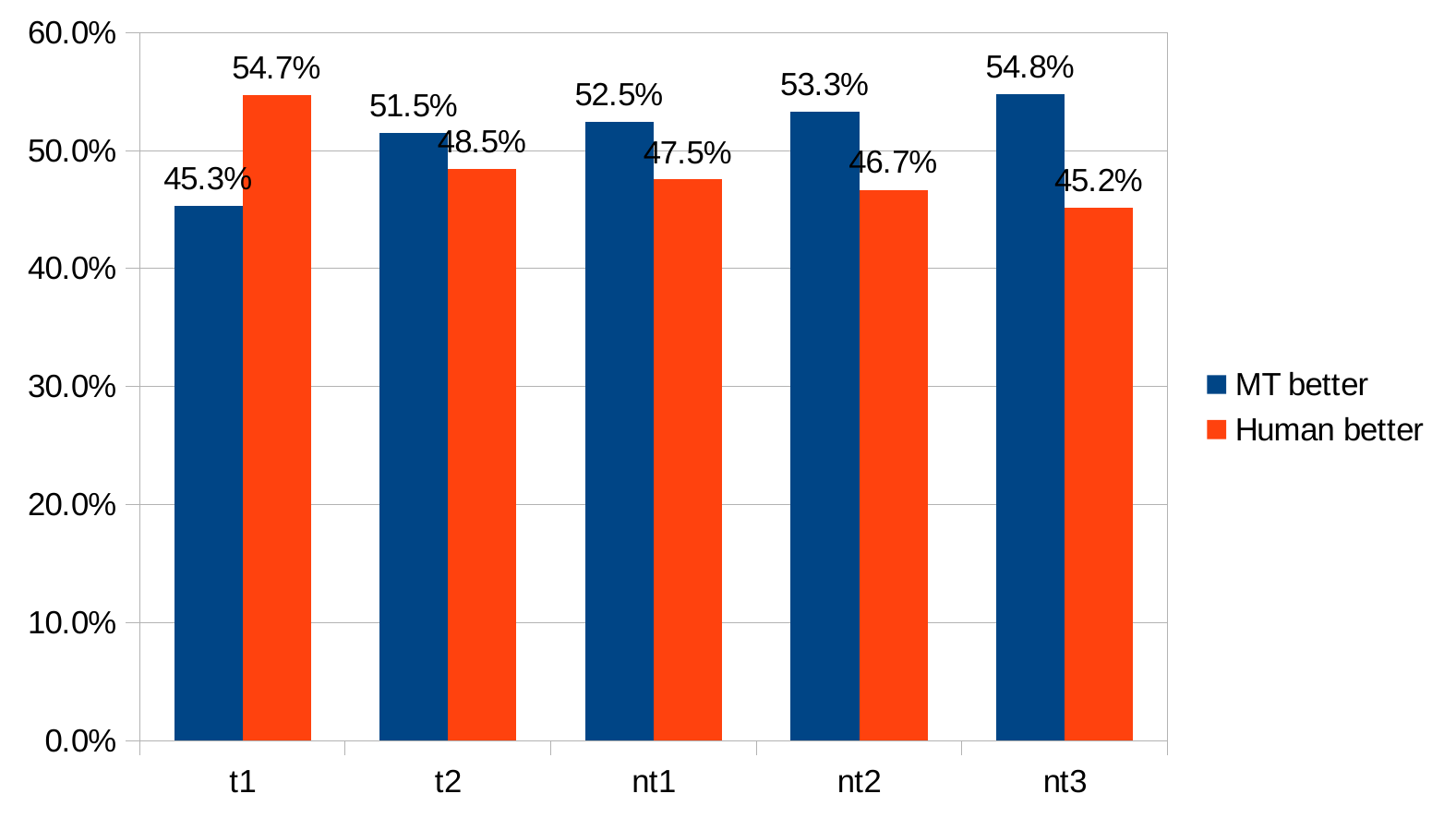}
  \caption{Results for English$\rightarrow$German for each evaluator separately: translators t1 and t2 and non-translators nt1, nt2 and nt3.}
  \label{f:ende_per_evaluator}
\end{figure}

Figure~\ref{f:ende_per_evaluator} shows the results for each evaluator separately, with ties omitted to ease the visualisation.
We observe a similar trend across all the non-translators: a slight preference for MT over \textsc{Human}, where the first is preferred in 52.5\% to 54.8\% of the times whereas the second is preferred in 45.2\% to 47.5\% of the cases.
However, the two translators do not share the same trend; translator t1 prefers \textsc{Human} more often than MT (54.7\% vs 45.3\%) while the trend is the opposite for translator t2, albeit more slightly (51.5\% MT vs 48.5\% \textsc{Human}).

\subsection{Results for English$\rightarrow$Russian}

\begin{figure}[htbp]
 \centering
 \includegraphics[width=0.5\textwidth]{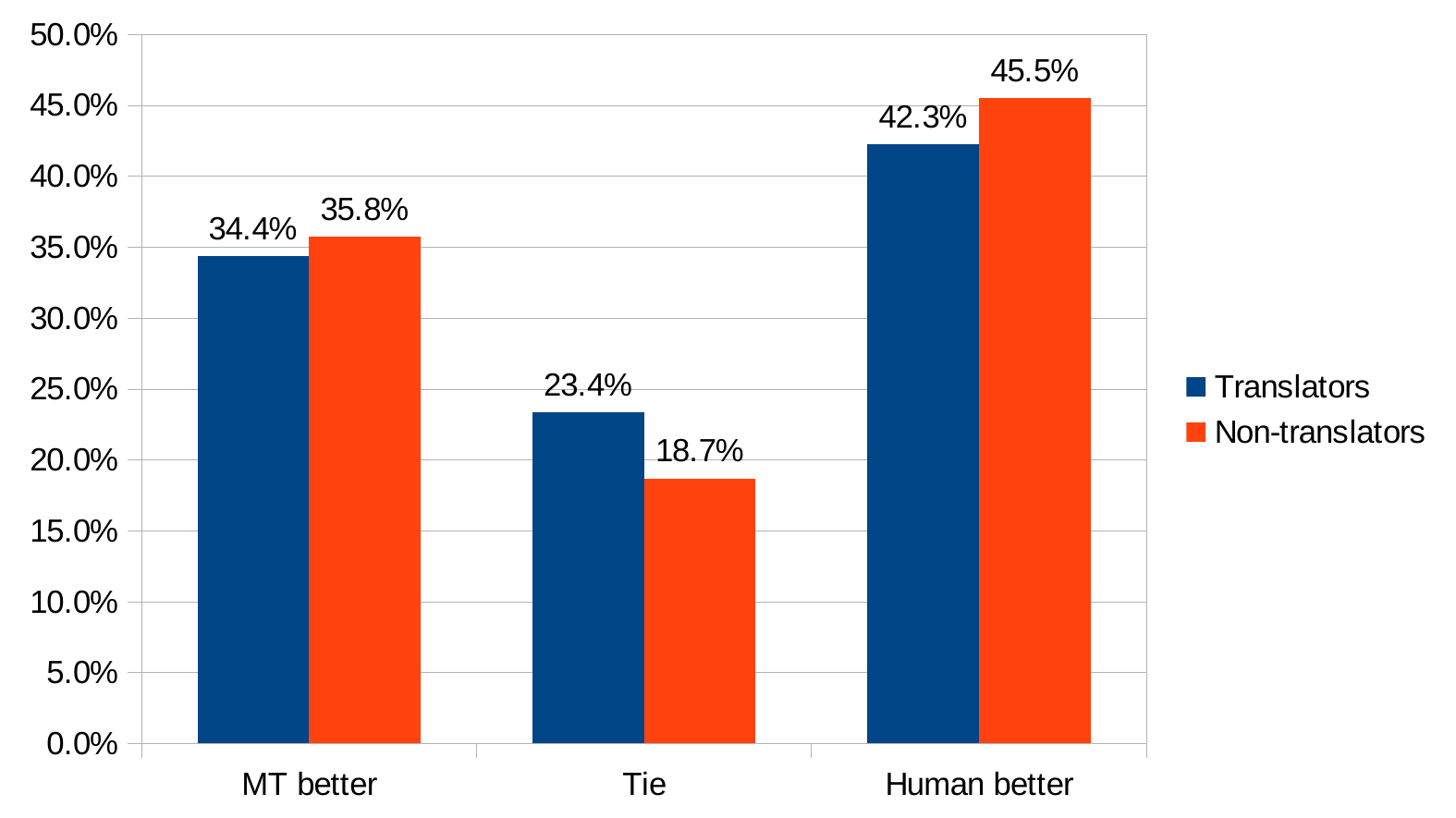}
  \caption{Results for English$\rightarrow$Russian for translators ($n=1181$) and non-translators ($n=604$)}
  \label{f:enru}
\end{figure}

Figure~\ref{f:enru} shows the results for English$\rightarrow$Russian.
In this translation direction both translators and non-translators prefer \textsc{Human} more frequently than MT: 42.3\% vs 34.4\% ($x=499$, $n=905$, $p<0.01$) and 45.5\% vs 35.8\% ($x=275$, $n=491$, $p<0.01$), respectively.
Since the differences are significant in both cases, our evaluation refutes the claim of human parity made at WMT 2019 for this translation direction.

\begin{figure}[htbp]
 \centering
 \includegraphics[width=0.5\textwidth]{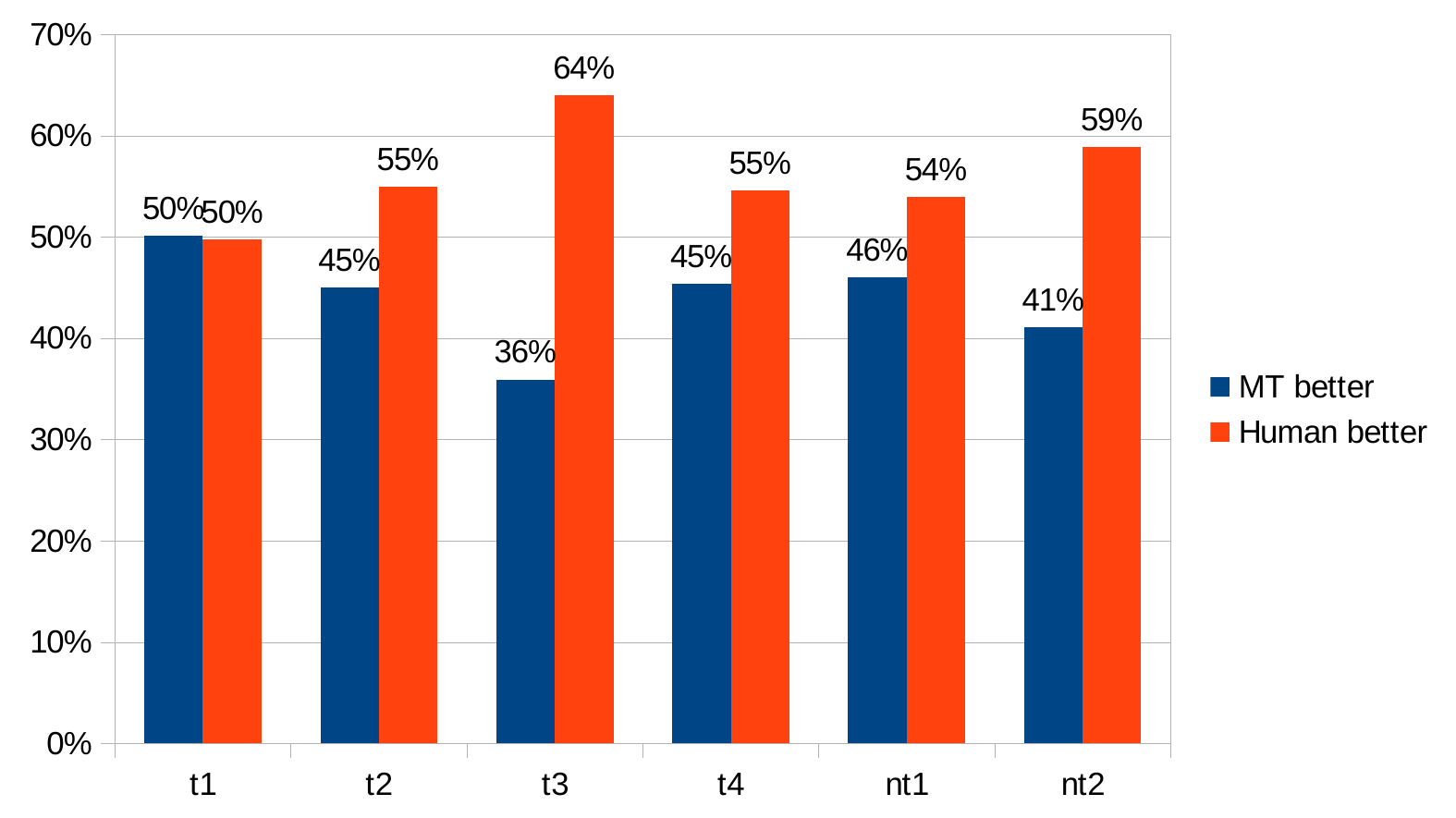}
  \caption{Results for English$\rightarrow$Russian for each evaluator separately: translators t1, t2, t3 and t4 and non-translators nt1 and nt2.}
  \label{f:enru_per_evaluator}
\end{figure}

Again we zoom in on the results by the individual evaluators, as depicted in Figure~\ref{f:enru_per_evaluator}.
It can be seen that all but one of the evaluators, translator t1, prefer \textsc{Human} considerably more often than MT.
However, the differences are only significant for t3 ($x=114$, $n=178$, $p<0.001$) and nt2 ($x=119$, $n=202$, $p<0.05$), probably due to the small number of observations.

\subsection{Results for German$\rightarrow$English}\label{s:res_deen}

\begin{figure}[htbp]
 \centering
 \includegraphics[width=0.5\textwidth]{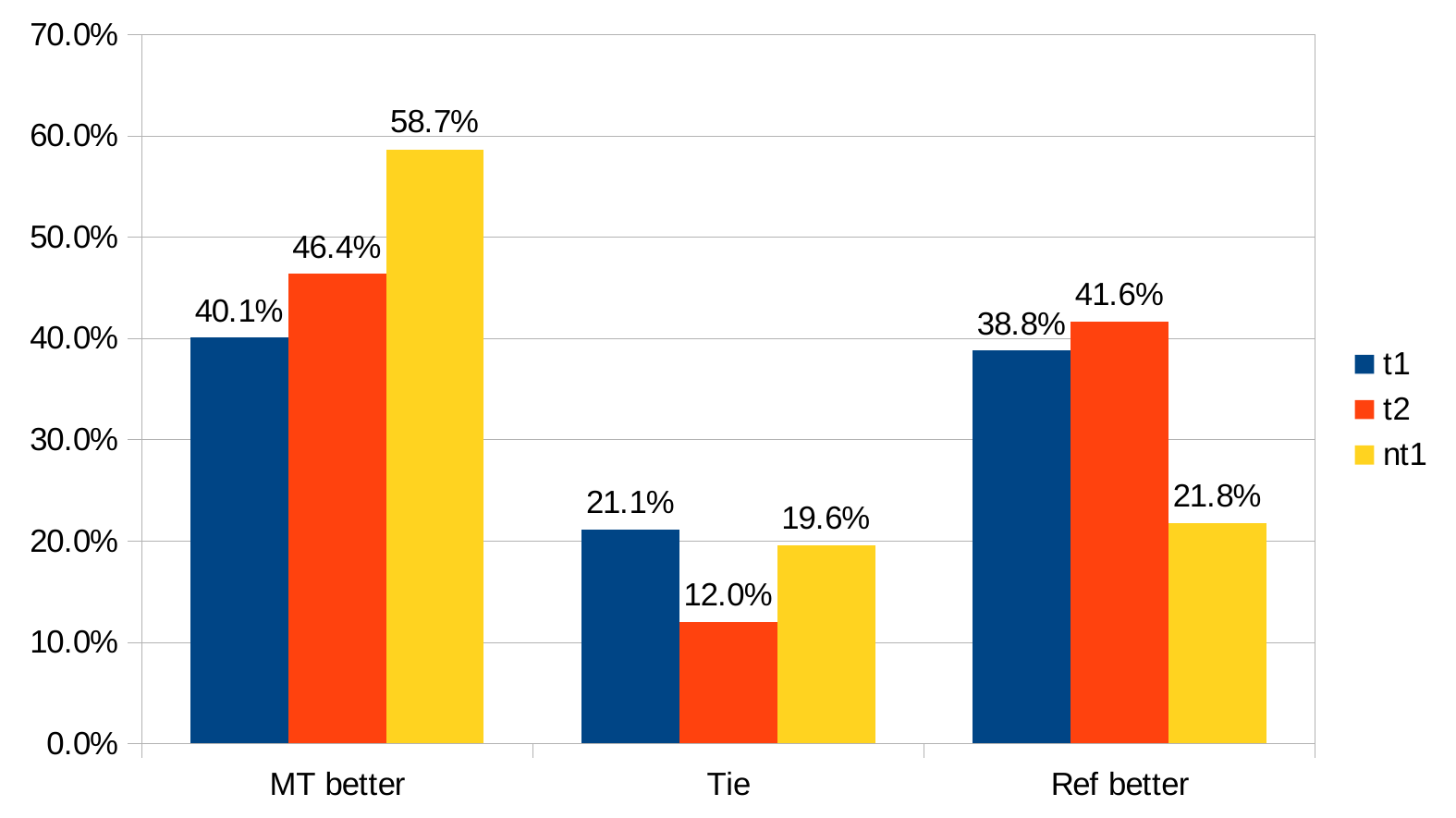}
  \caption{Results for German$\rightarrow$English for \textsc{Ref} and MT, with translators t1 and t2 and non-translator nt1.}
  \label{f:deen_mt_ref}
\end{figure}

\begin{figure}[htbp]
 \centering
 \includegraphics[width=0.5\textwidth]{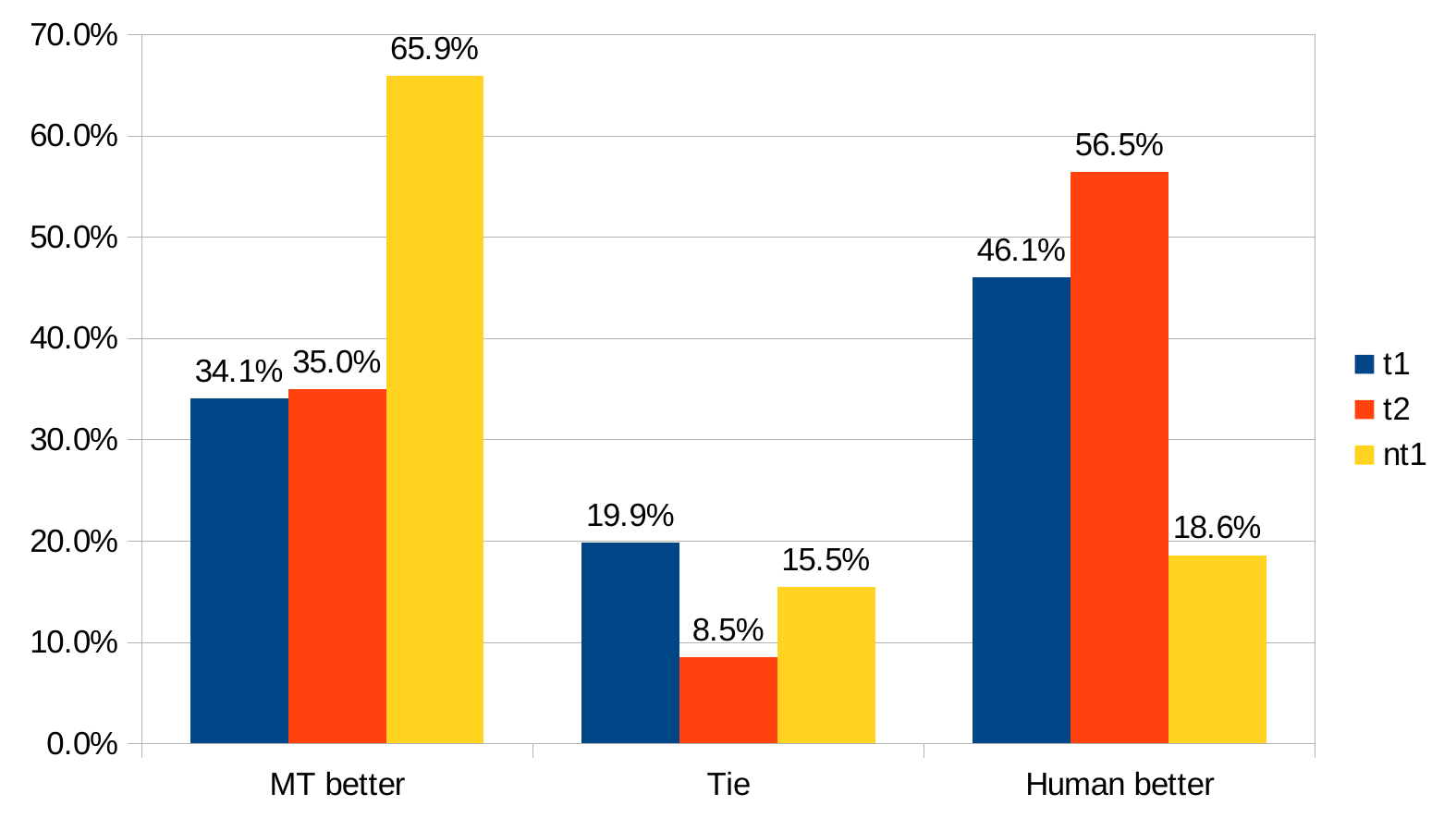}
  \caption{Results for German$\rightarrow$English for \textsc{Human} and MT, with translators t1 and t2 and non-translator nt1.}
  \label{f:deen_mt_human}
\end{figure}

\begin{figure}[htbp]
 \centering
 \includegraphics[width=0.5\textwidth]{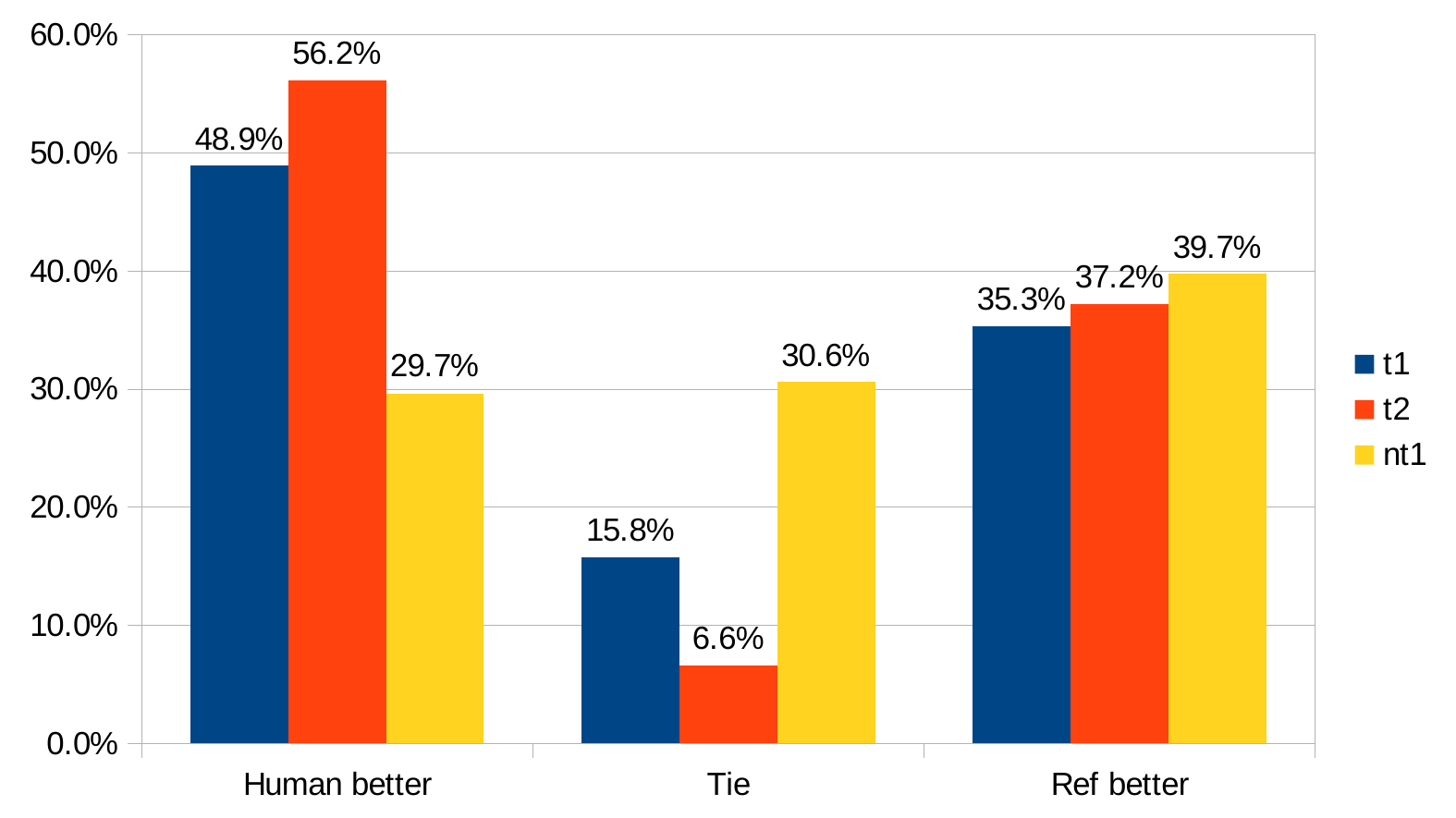}
  \caption{Results for German$\rightarrow$English for \textsc{Ref} and \textsc{Human}, with translators t1 and t2 and non-translator nt1.}
  \label{f:deen_human_ref}
\end{figure}

As explained in section~\ref{s:issues_refbased},
for this translation direction there are two human translations, referred to as \textsc{Human} and \textsc{Ref}, and one MT system.
Hence we can establish three pairwise comparisons: \textsc{Ref}--MT, \textsc{Human}--MT and \textsc{Human}--\textsc{Ref}.
The results for them are shown in Figure~\ref{f:deen_mt_ref}, Figure~\ref{f:deen_mt_human} and Figure~\ref{f:deen_human_ref}, respectively.

Both translators preferred the translation by the MT system slightly more often than the
human translation \textsc{Ref}, 
40\% vs 39\% and 46\% vs 42\%, 
but the difference is not significant ($x=255$, $n=529$, $p=0.4$).
The non-translator preferred the translation by MT considerably more often than \textsc{Ref}: 59\% vs 22\%, with the diffence being significant ($x=69$, $n=255$, $p<0.001$).
In other words, compared to \textsc{Ref}, the human translation used as gold standard at WMT 2019, the MT system achieves human parity according to the two translators and super-human performance according to the non-translator.

Now we discuss the results of comparing the MT system to the other human translation, \textsc{Human} (see Figure~\ref{f:deen_mt_human}).
The outcome according to the non-translator is, as in the previous comparison between \textsc{Ref} and MT, super-human performance ($x=59$, $n=268$, $p<0.001$), which can be expected since this evaluator prefers MT much more often than \textsc{Human}: 66\% vs 19\% of the times.
We expected that the results for the translators would also follow a similar trend to their outcome when they compared MT to the other human translation (\textsc{Ref}), i.e. human parity. However, we observe a clear preference for \textsc{Human} over MT: 
46\% vs 34\% and 57\% vs 35\%,
 resulting in a significant difference ($x=325$, $n=544$, $p<0.001$).

The last comparison is shown 
in Figure~\ref{f:deen_human_ref} and 
concerns the two human translations: \textsc{Ref} and \textsc{Human}.
The two translators exhibit a clear preference for \textsc{Human} over \textsc{Ref}: 
49\% vs 35\% and 56\% vs 37\%, 
($x=230$, $n=563$, $p<0.001$).
Conversely, the non-translator preferred \textsc{Ref} significantly  more often than \textsc{Human} ($x=126$, $n=220$, $p<0.05$): 40\% vs 30\%.

Given that
(i) parity was found between MT and \textsc{Human} in the reference-based evaluation of WMT, 
where \textsc{Ref} was the reference translation,
that (ii) \textsc{Human} is considerably different than \textsc{Ref} and MT (see Section~\ref{s:issues_refbased})
and that (iii) \textsc{Human} is found to be significantly better than \textsc{Ref} by translators in our evaluation,
it 
seems that reference bias played a role in the claim of parity at WMT. 

\subsection{Results of the Inter-annotator Agreement}

We now report the inter-annotator agreement (IAA) between the evaluators.
Since we have two types of evaluators, translators and non-translators, we report the IAA for both of them.
IAA is calculated in terms of Cohen's kappa coefficient ($\kappa$) 
as it was done at WMT 2016~\cite[Section~3.3]{bojar-EtAl:2016:WMT1}.

\begin{table}[htbp]
\centering
\begin{tabular}{c c c}
\hline
				       & \multicolumn{2}{c}{\textbf{Evaluators}}\\
				       \cline{2-3}
\textbf{Direction} & \textbf{ts}    & \textbf{nts} \\
\hline
English$\rightarrow$German & 0.326 & 0.266\\
English$\rightarrow$Russian & 0.239 & 0.238\\
German$\rightarrow$English & 0.320 & NA\\

\hline
\end{tabular}
\caption{Inter-annotator agreement with Cohen's $\kappa$ among translators (ts)
and non-translators (nts) for the three translation directions.}
\label{t:iaa}
\end{table}
Table~\ref{t:iaa} shows the IAA coefficients.
For English$\rightarrow$German, the IAA among translators ($\kappa=0.326$) is considerably higher, 23\% relative, than among non-translators ($\kappa=0.266$).
For English$\rightarrow$Russian, both types of evaluators agree at a very similar level ($\kappa=0.239$ and $\kappa=0.238$).
Finally, for German$\rightarrow$English, we cannot establish a direct comparison between the IAA of translators and non-translators, since there was only one non-translator.
However, we can compare the IAA of the two translators ($\kappa=0.32$) to that of each of the translators and the non-translator: $\kappa=0.107$ between the first translator and the non-translator and $\kappa=0.125$ between the second translator and the non-translator.
The agreement between translators is therefore 176\% higher than between one translator and the non-translator.

In a nutshell, for the three translation directions the IAA of translators is higher than, or equivalent to, that of non-translators, which corroborates previous findings by Toral et al.~\shortcite{toral-EtAl:2018:WMT}, where the IAA was 0.254 for translators and 0.13 for non-translators.

\subsection{Results of the Questionnaire}

The questionnaire (see Appendix~\ref{s:quest}) contained two 5-point Likert questions about how often additional context, local and global, was used.
In both cases, translators made slightly less use of context than non-translators: $M=2.9$, $SD=2.0$ versus $M=3.5$, $SD=1.0$ for local context and $M=1.4$, $SD=0.7$ versus $M=2$, $SD=0.9$ for global context.
Our interpretation is that translators felt more confident to rank the translations and thus used additional contextual information to a lesser extent.
If an evaluator used global context, they were asked to specify whether they used it mostly for some sentences in particular (those at the beginning, middle or at the end of the documents) or not.
Out of 8 respondents, 5 reported to have used global context mostly for sentences regardless of their position in the document
and the remaining 3 mostly for sentences at the beginning.

In terms of the perceived quality of the translations evaluated, all non-translators found one of the translations to be clearly better in general. Five out of the eight translators gave that reply too while the other three translators found all translations to be of similar quality (not so good).

Asked whether they thought the translations had been produced by MT systems or by humans, all evaluators replied that some were by humans and some by MT systems, except one translator, who thought that all the translations were by MT systems, and one non-translator who answered that he/she did not know.

\section{Conclusions and Future Work}\label{s:disc_fw}

We have conducted a modified evaluation on the MT systems that reached human parity or super-human performance at 
the news shared task of WMT 2019. 
According to our results: 
(i) for English$\rightarrow$German, the claim of super-human performance is refuted, but there is human parity;
(ii) for English$\rightarrow$Russian, the claim of human parity is refuted;
(iii) for German$\rightarrow$English, for which there were two human translations, the claim of human parity is refuted with respect to the best of the human translations, but not with respect to the worst. 


Based on our findings, we put forward a set of recommendations for human evaluation of MT in general and for the assessment of human parity in MT in particular:
\begin{enumerate}\itemsep0em
\item Global context (i.e. the whole document) should be available to the evaluator. Some of the evaluators have reported that they needed that information to conduct some of the rankings and contemporary research~\cite{castilho_lrec2020} has demonstrated that such knowledge is indeed required for the evaluation of some sentences.
\item If the evaluation is to be as accurate as possible then it should be conducted by professional translators. Our evaluation has corroborated  that evaluators that do not have translation proficiency evaluate MT systems more leniently than translators and that inter-annotator agreement is higher among the latter~\cite{toral-EtAl:2018:WMT}.
\item Reference-based human evaluation should be in principle avoided, given the reference bias issue~\cite{bentivogli2018machine}, which according to our results seems to have played a role in the claim of human parity for German$\rightarrow$English at WMT 2019.
That said, we note that there is also research that concludes that there is no evidence of reference bias~\cite{ma-etal-2017-investigation}.

\end{enumerate}

The first two recommendations were put forward recently~\cite{laubli2020set} and are corroborated by our findings. 
We acknowledge that our conclusions and recommendations are somewhat limited since they are based on a small number of sentences (just over 300 for each translation direction) and evaluators (14 in total).

Claims of human parity are of course not specific to translation.
Super-human performance has been reported to have been achieved in many other tasks, including board games, e.g. chess~\cite{Hsu:2002:BDB:601291} and Go~\cite{DBLP:journals/corr/abs-1712-01815}. 
However, we argue that assessing human parity in translation, and probably in other language-related tasks too, is not as straightforward as in other tasks such as board games, and that the former task poses, at least, two open questions, which we explore briefly in the following to close the paper.

\begin{enumerate}\itemsep0em
\item Against whom should the machine be evaluated?
In other words, should one claim human parity if the output of an MT system is perceived to be indistiguishable from that by an {\it average} professional translator or should we only compare to a {\it champion} professional translator?
In other tasks it is the latter case, e.g. chess in which \textsc{Deep Blue} outperformed world champion Gary Kasparov.
Related, we note that in tasks such as chess it is straightforward to define the concept of a player being better than another: whoever wins more games, the rules of which are deterministic.
But in the case of translation, 
it is not so straightforward to define whether a translator is better than another.
This question is pertinent since, as we have seen for German$\rightarrow$English (Section~\ref{s:res_deen}), where we had translations by two professional translators, the choice of which one is used to evaluate an MT system against can lead to a claim of human parity or not.
In addition, the reason why one claim remains after our evaluation (human parity for English$\rightarrow$German) might be that the human translation therein is not \emph{as good as it could be}. 
Therefore, once the three potential issues that we have put forward (see Section~\ref{s:issues}) are solved, we think that an important potential issue that should be studied, and which we have not considered, has to do with the quality of the human translation used. 

\item Who should assess claims of human parity and super-human performance?
Taking again the example of chess, this is straightforward since one can just count how many games each contestant (machine and human) wins.
In translation, however, we need a person with knowledge of both languages to assess the translations. We have seen that the outcome is dependent to some extent on the level of translation proficiency of the evaluator: it is more difficult to find human parity if the translations are evaluated by professional translators than if the evaluation is carried out by  bilingual speakers without any translation proficiency.
Taking into account that most of the users of MT systems are not translators, should we in practice consider human parity if those users do not perceive a significant difference between human and machine translations, even if an experienced professional translator does?


\end{enumerate}

\section*{Acknowledgments}
This research has received funding from CLCG's 2019 budget for research participants.
I am grateful for valuable comments from Barry Haddow, co-organiser of WMT 2019. I would also like to thank the reviewers; their comments have definitely led to improve this paper.

\bibliography{wmt19_parity}
\bibliographystyle{eamt20}

\appendix
\section{Post-experiment Questionnaire}\label{s:quest}

{\small

\begin{enumerate}\itemsep0em
\item Rate your knowledge of the source language
\begin{itemize}
	\item None; A1; A2; B1; B2; C1; C2; native
\end{itemize}
\item Rate your knowledge of the target language
\begin{itemize}
	\item None; A1; A2; B1; B2; C1; C2; native
\end{itemize}

\item How much experience do you have translating from the source to the target language?
\begin{itemize}
	\item None, and I am not a translator; None, but I am a translator; Less than 1 year; between 1 and 2 years; between 2 and 5 years; more than 5 years
\end{itemize}

\item During the experiment, how often did you use the local context shown in the web application (i.e. source sentences immediately preceding and immediately following the current sentence)?
\begin{itemize}
	\item Never; rarely; sometimes; often; always
\end{itemize}

\item During the experiment, how often did you use the global context provided (i.e. the whole source document provided as a text file)?
\begin{itemize}
	\item Never; rarely; sometimes; often; always
\end{itemize}

\item If you used the global context, was that the case for ranking some sentences in particular?
\begin{itemize}\itemsep0em
	\item Yes, mainly those at the beginning of documents, e.g. headlines
	\item Yes, mainly those in the middle of documents
	\item Yes, mainly those at the end of documents
	\item No, I used the global context regardless of the position of the sentences to be ranked
\end{itemize}

\item About the translations you ranked
\begin{itemize}\itemsep0em
	\item Normally one was clearly better
	\item All were of similar quality, and they were not so good
	\item All were of similar quality, and they were very good
\end{itemize}

\item The translations that you evaluated were in your opinion:
\begin{itemize}\itemsep0em
	\item All produced by human translators
	\item All produced by machine translation systems
	\item Some produced by humans and some by machine translation systems
	\item I don't know
\end{itemize}

\end{enumerate}
}

\end{document}